\pdfoutput=1

\documentclass[11pt]{article}

\usepackage{ACL2023}

\usepackage{graphicx}
\usepackage{tikz}
\usetikzlibrary{shapes, arrows, positioning}

\usepackage{subfig}
\usepackage{booktabs, tabularx}
\usepackage[utf8]{inputenc} 
\usepackage[T1]{fontenc}    
\usepackage[english,german]{babel}  
\usepackage{hyperref}       
\usepackage{url}            
\usepackage{amsfonts}       
\usepackage{nicefrac}       
\usepackage{microtype}      
\usepackage{xcolor}         
\usepackage{wrapfig}
\usepackage{times}
\usepackage{latexsym}
\usepackage[ruled,vlined,noend]{algorithm2e} 

\usepackage{pgfplots}
\usepackage{enumitem}
\usepackage{diagbox}
\usepackage{tkz-tab}
\usepackage{caption}
\usepackage{amssymb}
\usepackage{amsmath}
\usepackage{subcaption}
\usepackage{natbib}
\usepackage{blindtext}
\usepackage{ntheorem}
\usepackage{tfrupee}

\SetAlFnt{\small}
\SetAlCapFnt{\small}
\SetAlCapNameFnt{\small}

\DeclareUnicodeCharacter{200B}{}

\usepackage[T1]{fontenc}

\usepackage[utf8]{inputenc}

\usepackage{microtype}

\usepackage{inconsolata}

\title{On Instruction-Finetuning Neural Machine Translation Models}

\author{%
Vikas Raunak \quad Roman Grundkiewicz \quad Marcin Junczys-Dowmunt\\
Microsoft Azure AI\\
\texttt{\{viraunak,rogrundk,marcinjd\}@microsoft.com}\\}

\begin{document}
\maketitle
\begin{abstract}

In this work, we introduce instruction finetuning for Neural Machine Translation (NMT) models, which distills instruction following capabilities \textit{from} Large Language Models (LLMs) \textit{into} orders-of-magnitude smaller NMT models. 
Our instruction-finetuning recipe for NMT models enables customization of translations for a limited but disparate set of translation-specific tasks.
We show that NMT models are capable of following multiple instructions simultaneously and demonstrate capabilities of zero-shot composition of instructions.
We also show that through instruction finetuning, traditionally disparate tasks such as formality-controlled machine translation, multi-domain adaptation as well as multi-modal translations can be tackled jointly by a single instruction finetuned NMT model, at a performance level comparable to LLMs such as GPT-3.5-Turbo.
To the best of our knowledge, our work is among the first to demonstrate the instruction-following capabilities of traditional NMT models, which allows for faster, cheaper and more efficient serving of customized translations.


\end{abstract}

\section{Introduction}

Instruction-finetuned Large Language Models (LLMs) demonstrate the remarkable ability of instruction-following \citep{jason_ift_few_shot_learners}, which makes them amenable to tackle any task cast as natural language generation, even under a zero-shot setting.
In this work, we explore whether traditional Neural Machine Translation (NMT) models could offer \textit{similar} capabilities of following instructions. NMT models could be considered as domain-specific `language' models
\textit{pre-trained} for a single task (translation) and thereby \textit{could} be instruction-finetuned to tackle translation-adjacent tasks such as translation customization or enforcing certain specifications on the translations.
Such tasks, e.g., formality-controlled translation \cite{additive_interventions}, multi-modal translation \cite{elliott-etal-2016-multi30k} or gender-based translation rewriting \cite{kuczmarski2018gender}, have typically been tackled through specialized models or algorithms in prior literature, rather than a single instruction-following NMT model. In contrast, we instruction-finetune a single \textit{ancestral} translation model to \textit{adapt} the translations based on instructions. Our contributions are as follows:

\begin{enumerate}
    \item We present a new recipe for instruction finetuning NMT models (trained with supervision only on parallel datasets), which allows for joint modeling of disparate translation customization tasks in a single NMT model, and we analyze the criticality of each of the recipe components through ablation experiments.
    \item We demonstrate that NMT models are capable of following multiple (30+) instructions simultaneously. We also find that NMT models show abilities of zero-shot composition of instructions, as an effect of finetuning.
    \item We show that, with a single instruction-finetuned NMT model, traditional customization tasks such as formality-controlled machine translation can be tackled with high performance, in conjunction with several disparate tasks.
\end{enumerate}

\noindent
Additionally, our proposed finetuned NMT model outperforms GPT-3.5-Turbo on average on the IWSLT-22 Formality Control Shared Task \cite{iwslt2022}, while simultaneously achieving high-performance on others \& demonstrating a few other \textit{desirable} properties vis-à-vis much larger LLMs. At a high-level, our work re-interprets a NMT model as a language model and demonstrates the utility of instruction finetuning NMT model for jointly modeling a myriad of disparate translation-related tasks.
In the next sections, we elaborate on our recipe for instruction-finetuning of a NMT model and analyze its characteristics.

\begin{table*}[ht]
    \small
    \renewcommand{\arraystretch}{1.4} 
    \setlength{\tabcolsep}{8pt} 
    
    \begin{tabularx}{\textwidth}{|>{\hsize=.33\hsize}X|>{\hsize=.33\hsize}X|>{\hsize=.33\hsize}X|}
        \hline
        \textbf{Instruction Prefix} & \textbf{Source (English)} & \textbf{Translation (German)} \\ 
        \hline \hline

        \emph{past tense} & The finished effect \colorbox{orange}{is} long-lasting and highly glossy – but does it damage the nails? & Der fertige Effekt \colorbox{orange}{war} langanhaltend und hochglänzend – aber beschädigte er die Nägel? \\ \hline

        \emph{informal} & Do you like Legos? did you ever play with them as a child or even later? & \colorbox{orange}{Magst du} Legosteine? \colorbox{orange}{Hast du} jemals als Kind oder sogar später mit ihnen gespielt?​ \\ \hline

        \emph{fix misspelling} & To switch between environments, update the storage.json file with the URL of the \colorbox{orange}{specificrrbzpronment}. & Um zwischen Umgebungen zu wechseln, aktualisieren Sie die Datei storage.json mit der URL des \colorbox{orange}{spezifischen Prozesses}. \\ \hline

        \emph{translate "herbal medicines" to "Kräutermedizin"} & Chinese \colorbox{orange}{herbal medicines} for hypothyroidism & Chinesische \colorbox{orange}{Kräutermedizin} gegen Hypothyreose \\ \hline

        
        \centering\includegraphics[width=0.2\textwidth]{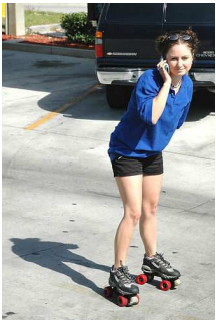} & A trendy girl talking on her cellphone while gliding slowly down the street. & Ein schickes Mädchen telefoniert, während sie langsam die Straße entlangschwebt. \\ 
        \hline
    \end{tabularx}
    
    \caption{Input-output instances for the developed instruction finetuned NMT model. 
    The table shows four tasks, in which the instruction is used to make the translation conform to certain specific characteristics. 
    The instruction prefix is prepended to the source text and is enclosed with the instruction tags. 
    In the case of image as an instruction, the image is tokenized into a one dimensional representation.}
    \label{tab:inst_examples}
    \vspace{-1.0em}
\end{table*}


\section{Related Work}

Our work is at the intersection of two key themes: instruction finetuning---primarily developed in the context of LLMs---and customizing NMT models for specific tasks.

\subsection{Instruction Finetuning of LLMs}

Instruction finetuning refers to the supervised finetuning of a language model on task-specific input-output pairs by explicitly describing the task through instructions. 
This has been demonstrated to aid in cross-task generalization \cite{t0, flan_collection}, in particular, imparting LLMs with instruction-following capabilities \cite{jason_ift_few_shot_learners}. 
A number of prior works have proposed different algorithms for constructing the instruction data \cite{ift1, ift2, ift3, ift4, ift5, ift6, ift7, ift8}.

In our recipe, we rely on a combination of parallel data filtering and synthetic data generation through LLMs to construct the instruction dataset that is leveraged for finetuning NMT models. 
Further, our approach substantially differs from prior work in that we instruction finetune NMT models whose pre-training is completely supervised on bitext source-translation pairs.

\subsection{Customizing Translation Models}

There exists a large body of work in adapting NMT models and customizing them for specific use cases such as for achieving high-performance on specific domains \cite{saunders2022domainadaptationmultidomainadaptation}, tones or registers in the target language \cite{nădejde2022cocoamtdatasetbenchmarkcontrastive} as well as for tasks such as gender-based translation rewriting \cite{rarrick2023gatechallengesetgenderambiguous}. 
Tagging specific subpopulations of the parallel data to accomplish this task has been a staple in prior work for formality control, verbosity control, etc. 

Our work is related to the tagging approaches developed in the literature but differs in two key aspects: 
(a) task diversity and scale: typically, tagging is only applied to supply information pertaining to a single task, while instruction finetuning as a technique aspires to tackle a wide variety of tasks in a unified modeling approach to make the model capable of following a wide variety of instructions;
and (b) natural language instruction: instead of manipulating tags or combination of tags, we leverage instructions expressed or composed in natural language for influencing the translations. 

\section{Instruction Finetuning of NMT models}
\label{gen_inst}

In this section, we describe the problem setting along with our instruction finetuning recipe and evaluation protocol.

\subsection{Problem Setting} 

For instruction finetuning, we take a pre-trained NMT model and finetune it with instruction annotated source-translation pairs. 
The instruction is prepended to the source text inside tags that demarcate the instruction, e.g., \textit{<instruction> informal </instruction>}. 
Henceforth, we refer to the tokens pertaining to the \textit{<instruction>} and \textit{</instruction>} strings as the instruction tokens. 
A collection of instruction and source-translation instances are presented in Table \ref{tab:inst_examples}. 
Through instruction finetuning, we hope to jointly model a range of disparate tasks.

\subsection{Instruction Finetuning Recipe}
\label{ssec:layout}

\begin{algorithm}[t]
    \SetAlgoNoLine
    \SetNoFillComment

    \vspace{1em} 
    
    \KwData{Base NMT Model and Vocabulary}
    \KwResult{Instruction Finetuned NMT Model}
    
    \vspace{1em} 
    
    \textbf{Step 1:} \quad Expand vocabulary with instruction tokens \\
    \vspace{0.5em} 

    \textbf{Step 2:} \quad Curate task-specific and parallel datasets \\
    \vspace{0.5em} 
    
    \textbf{Step 3:} \quad Finetune on a \textit{mix} of parallel and task data \\
    \vspace{0.5em} 
    
    \textbf{Step 4:} \quad (Optional) Interpolation with base model
    
    \vspace{1em} 

    \caption{Instruction-Finetuning NMT Recipe}
    \label{algo:PPA}
\end{algorithm}

We present our simple recipe for instruction finetuning NMT models in Algorithm \ref{algo:PPA}.
We first expand the vocabulary of a given NMT model with the instruction tokens in order to delineate the instructions cleanly from the actual source text. 
Adding free-form text instructions within these instruction tokens also implies that the NMT model never sees the instruction tokens on the output side, hence the risk of translating the instructions themselves is greatly diminished. 
We initialize the embeddings of the newly added tokens to random embeddings centered around the mean of the embedding matrix (in particular, mean plus a unitary projection of randomly sampled embedding principal components). 

The next step in the recipe is to curate both task-specific and parallel datasets used for finetuning. 
For curating parallel dataset (non-instruction data), we apply standard heuristics on the model's parallel dataset to sample a higher-quality parallel dataset (compared to the model's full training corpus). The details of the heuristics are presented in Appendix \ref{sec:appendix_d}.
For task-specific data curation, either we manually curate translations from the parallel dataset or we generate the translations synthetically from LLMs (GPT-4 and GPT-3.5-Turbo). 
We describe task specific dataset curation in section 3.4. 

Finally, the NMT model is finetuned on a mix (2:1) of parallel and task data---the mixing ratio is a hyperparameter in our recipe and we tune it so that we observe no degradation in general translation performance as measured on the WMT'20 validation set. 
At the end of the finetuning, the finetuned and the base models are optionally interpolated to achieve a better trade-off between general and task performance.
We present the details of the interpolation step in the Appendix \ref{sec:appendix_a}, while the details pertaining to the other steps are presented in the next sections. 
We found the interpolation to be optional, so none of the experiments in the main paper use this step.

\subsection{Evaluation Protocol}
\label{ssec:eval_prot}

For the instruction finetuned NMT model, we have the choice of either translating an input without any instruction (the \textit{general} case) or using a particular instruction (the \textit{instruction} case). 
Throughout this work, we report the following measurements in order to evaluate the instruction finetuned NMT model:
\begin{enumerate}

    \item \textbf{General Performance}: This is measured by computing the MT quality of the finetuned NMT model (i.e., the original translation task) on a standard test set. 
    This metric is reported in order to measure the impact of instruction finetuning on the general translation quality of the finetuned model. 
    
    \item \textbf{Task-Specific Performance}: On a per-task basis we report two measurements:
    \begin{enumerate}[label=\alph*.]
        \item  \textbf{Task Response Rate (RR)}: the percentage of instances in the test set for which including a instruction yielded a different translation than not including the instruction (the \textit{general} case). 
        This offers us a crude measure to evaluate how responsive the model is to a specific instruction.
        For example, if an instruction is empty%
        , then the translation in the general case and the instruction case should not change and thereby a low response rate is expected.
        \item \textbf{Task Output Quality}: the MT quality metrics (over system outputs and references) for the finetuned NMT model both in the \textit{general} case and the \textit{instruction} case. 
        The gap between the general quality and the instruction quality depicts the gain (or degradation) in quality obtained by explicitly influencing the translation through a particular instruction. 
    \end{enumerate}
        
\end{enumerate}

Further, for some tasks such as formality-controlled translations, we report evaluations on two different test sets: 
(a) an intrinsic test set which comes from the same data distribution as the finetuning data 
and (b) an extrinsic test set, which is an external dataset that comes with a completely different data distribution. 
Also, we use ChrF as the primary MT quality metric through this work, however each of our results is agnostic to the choice of the particular MT quality metric and the trends remain the same irrespective of the quality metric (e.g., COMET) used.

\section{Experiments} 

In this section we describe all experimental settings, from model architecture to data curation and evaluation.

\subsection{Experimental Settings}

We conduct experiments on the WMT'20 News Translation (English-German) task benchmark \cite{wmt-2020-findings}. 
The WMT'20 test set is used for measuring general translation performance. 
We used the official parallel training data from WMT'20 with the dataset statistics presented in Table \ref{tab:data_source}. 
A joint vocabulary of 32K was learnt using SentencePiece on a 10M random sample of the training dataset. 

The trained model is a Transformer-Big (225M parameters) with the hyperparameters described exactly in \citet{transformer}. 
The model was trained for 300K updates using Marian NMT \cite{marian}. 
The metrics BLEU, ChrF2, TER \cite{bleu, chrf, ter} for the trained model on the WMT'20 validation and test sets (under beam size of $1$) as measured using SacreBLEU \cite{sacrebleu} are presented in Appendix \ref{sec:appendix_b}, alongside reference-based COMET \cite{comet} scores.

\begin{table}[ht!]
    \centering
    \setlength\tabcolsep{8.5pt}
    \begin{tabular}{lr}
    \toprule 
\textbf{Data Source} &  \textbf{Sentence Pairs}    \\
        \midrule
Europarl  & 1,828,521    \\
ParaCrawl &	34,371,306  \\
Common Crawl &	2,399,123 \\
News Commentary	& 361,445 \\
Wiki Titles	& 1,382,625 \\
Tilde Rapid	& 1,631,639 \\
WikiMatrix &	6,227,188 \\ \midrule
Total  & 48,201,847	\\
        \bottomrule
    \end{tabular}
    \caption{The WMT'20 data sources used for training the English--German NMT model.}
    \label{tab:data_source}
\end{table}



\begin{table*}
  \centering
  \begin{tabular}{lllll}
    \toprule
    Task Instruction     & RR (\%)  & ChrF$_{\text{general}}$  & ChrF$_{\text{instruction}}$  & Improvement \\
    \midrule
    past tense & 84.81  & 82.06  & 86.85 & + 4.79  \\
    translate X to Y     & 60.42 & 76.18 &  80.24 & + 4.06     \\
    active voice     &  54.84 & 87.62  & 92.86 & + 5.24 \\
    passive voice     & 80.91 & 71.44 & 78.29 & + 6.85 \\
    non-literal & 50.00 & 83.25 & 84.89 & + 1.64 \\
    literal & 53.41 & 90.12 & 92.88 & + 2.76 \\
    titlecase & 100.0  & 52.75  & 68.52 & + 15.77 \\
    lowercase     & 100.0 & 55.39 & 67.35 & + 11.96    \\
    uppercase     & 98.92 & 2.41 & 40.31 & + 37.9 \\
    remove punctuation & 100.0 & 67.18 & 68.73 & + 1.55 \\
    add antonyms & 79.79  & 71.90  & 73.12 & + 1.22 \\
    remove profanity     &  66.67 & 75.81   & 77.38 & + 1.57  \\
    add hashtag & 100.0 & 61.05 & 68.68 & + 7.63  \\
    leetify & 100.0 & 26.37 & 34.12 & + 7.75 \\
    remove accents & 81.97 & 59.55 & 62.08 & + 2.53\\
    shuffle words & 100.0 & 52.69 & 42.62 & - 10.07 \\
    fix misspelling & 91.74 & 60.22 & 65.36 & + 5.14 \\
    introduce repetition error & 55.34 & 64.54 & 65.36 & + 0.82  \\
    insert X at the beginning & 100.0 & 64.78 & 69.19 & + 4.41 \\
    insert X at the end & 100.0 & 64.38 & 69.68 & + 5.3 \\
    same length & 58.16 & 89.37 & 95.93 & + 6.56 \\
    shorter length & 52.59 & 90.88 & 94.30 & + 3.42 \\
    longer length & 57.38 & 66.51 & 68.14 & + 1.63 \\
    simplify & 81.42 & 61.88 & 67.22 & + 5.34 \\
    complexify & 58.33 & 89.31 & 93.92 & + 4.61 \\
    obsfuscate & 56.84 & 80.89 & 82.61 & + 1.72 \\
    formal & 60.77 & 86.53 & 91.03 & + 4.50  \\
    informal & 60.58 & 87.28 & 92.25 & + 4.97 \\
    spacing error & 84.40 & 66.70 & 66.87 & + 0.17 \\ 
    coverage error & 97.25 & 66.40 & 66.24 & - 0.16 \\  
    image (multi-30k) & 53.00 & 72.08 & 74.89 & + 2.81 \\
    empty instruction & 0.06 & 65.27 & 65.27 & + 0.0 \\ \midrule
    average & 89.60 & 74.20 & 82.42 & + 8.22 \\
    \bottomrule \\
  \end{tabular}
  \vspace{-0.3cm}
  \caption{Intrinsic evaluation results for the instruction finetuned NMT system over different tasks. 
  Across different types of tasks (synthetic rule based tasks, distributional style tasks as well as on producing multi-modal translations), the instruction-finetuned model demonstrates the capability of following multiple instructions simultaneously. Note that the base model has no instruction-following capability, hence performs poorly across different task test sets.}
  \label{tab:intrinsic}
\end{table*}


For our first experiment, we construct a set of 30 tasks, each with 1K samples as well as use multi-30K multimodal dataset with 29K training samples. 
For multi-30K, we convert the image into 32 tokens using 1D image tokenizer\footnote{\url{https://github.com/bytedance/1d-tokenizer}} from \citet{yu2024imageworth32tokens}. 
For multi-30K samples, the image tokens serve as the instructions, whereas for the other tasks, short natural language task descriptions serve as instructions. 
Further details for these tasks are presented in Appendix \ref{sec:appendix_c}. 
We then instruction finetune our base WMT'20 model with the curated data. 
Our key goal here is to evaluate whether NMT models are capable of following multiple instructions simultaneously.

\subsection{Task-Specific Data Curation}

The first column of Table \ref{tab:intrinsic} shows the list of task instructions. 
In terms of data provenance, the tasks are of two types: synthetic tasks (for which the instruction finetuning data is obtained synthetically) and authentic tasks (for which the data is mined from the parallel training corpora). 
We present a more verbose description of each of the tasks in Appendix \ref{sec:appendix_c}, since the text in the instruction naturally implies the targeted translation task. 

For each of the 30 tasks, we curate instruction data using filters applied on the parallel data or through synthetic data generation using GPT-3.5-Turbo or GPT-4.
In particular, the data for instructions pertaining to generating active voice, passive voice, simplifying, complexifying and obsfuscating translations were obtained synthetically through GPT-3.5-Turbo\footnote{ \url{https://beta.openai.com/docs/models/}}, whereas formal and informal translation data was obtained using GPT-4.

\begin{table*}[ht]
  \centering
  \begin{tabular}{llllll}
    \toprule
    Task Instruction     & RR (\%)  & ChrF$_{\text{general}}$  & ChrF$_{\text{instruction}}$ & T$_1$ SR (\%) & T$_2$ SR (\%) \\
    \midrule
    lowercase & 100.00  & 53.82  & 68.11 & 83.00 &  --    \\
    uppercase & 100.00  & 2.42  & 44.67 & 27.96 &  -- \\ \midrule
    remove profanity     & 93.33 & 69.88 & 80.95 & -- &  40.00    \\ \midrule
    lowercase remove profanity     & 100.00 & 58.86 & 70.69 & 80.00 & 40.00  \\
    uppercase remove profanity & 100.00 & 2.97 & 39.31 & 26.67 & 6.67 \\ \midrule
    lowercase and remove profanity     & 100.00 & 58.86 & 69.23 & 93.33 & 33.33  \\
    uppercase and remove profanity & 100.00 & 2.97 & 43.27 & 26.67 & 13.33 \\
    \bottomrule \\
  \end{tabular}
  \vspace{-0.3cm}
  \caption{Zero-shot composition of instructions. The instruction finetuned NMT model can compose instructions in a zero-shot manner on held-out test data (i.e., the model has not been trained on any combinations of instructions).
  Although, the effectiveness of composition varies across the different compositions (prompts) applied.
  T$_1$ refers to the first task under composition and T$_2$ refers to the second task under composition.}
  \label{tab:zeroshot}
\end{table*}

\begin{table*}
  \centering
  \begin{tabular}{lcc}
    \toprule
    Formality-Control Translation Model     & Formal Accuracy  & Informal Accuracy  \\
    \midrule
    mBART-large, \citet{rippeth-etal-2022-controlling} &  93.6 &  77.4   \\
    LLM, \citet{garcia2023unreasonableeffectivenessfewshotlearning}   & 84.9 &  85.5     \\
    Doc-MT System, \citet{post2024escapingsentencelevelparadigmmachine}    & 83.3  & 87.1   \\
    GPT-3.5-Turbo\footnote{https://platform.openai.com/docs/models/gpt-3-5-turbo} & 95.5 & 95.0   \\ \midrule
    (ours) Baseline WMT-20 model &  75.0 & 25.0   \\
    (ours) Instruction-Finetuned WMT-20 model   & 94.7 &  98.5    \\ \midrule
    WMT'22 Task Winner (Constrained)     & 100.0 & 88.6 \\
    WMT'22 Task Winner (Unconstrained) & 100.0 & 100.0  \\
    \bottomrule
  \end{tabular}
\caption{Extrinsic evaluation on  producing formal and informal translations.
  The instruction finetuned NMT model outperforms GPT-3.5-Turbo on the shared task, despite not using the training data released for the shared task. 
  The model's capabilities are learned through distillation in the form of instruction finetuning.}
  \label{tab:extrinsic}
\end{table*}

\subsection{Finetuning and Evaluation Settings} 

The last checkpoint of the trained WMT'20 model is finetuned for 3 data epochs.
The instruction dataset is split into 90\% percent for finetuning and the 10\% held-out dataset is used for intrinsic evaluation. 
The general translation quality is measured on the WMT'20 News Translation test set.


\section{Results and Analysis}
\label{sec:results}

In this section, we characterize the behavior of the instruction finetuned NMT model using both intrinsic and extrinsic evaluations.
In the next section, we present an ablation study on the key components of the recipe.

\subsection{Instruction-Following Performance}

Table \ref{tab:intrinsic} presents the results that characterize the instruction-following performance of the finetuned NMT model.
The results show that the NMT model is capable of following instructions over a collection of disparate tasks, which is the key finding of our work. 

In particular, both rule-based tasks such as \emph{leetify} (which inserts leet-speak in the translation) as well as tasks which are more distributional and style based in nature, such as \emph{complexify}, are remarkably well learned by the NMT model. 
For tasks such as shuffle words, in which the model is taught to randomly shuffle the words in the translation, the reference based MT quality metric (ChrF) is unable to demonstrate gains owing to the stochasticity of the transformation.

\subsection{Zero-Shot Composition of Instructions}

Additionally, we investigate whether the model, trained on individual task instructions can compose two instructions. 
Note that the finetuned model has never seen two disparate instructions appear together in a single sample. 
We find that the model is capable of composing instructions in a zero-shot manner and Table \ref{tab:zeroshot} presents an example of such a composition. 

To further investigate this behavior, in Table \ref{tab:zeroshot}, we present additional metric named Task Success Rate (SR), which provides a binary measure of the task success rather than a continuous measure such as ChrF. 
Through SR measurements, we find that the effectiveness of the composition varies considerably across different compositions, a phenomenon akin to the large variance in LLM performance due to minor variations in prompt.

\subsection{Extrinsic Evaluations}
\label{ssec:ext_eval}

We conduct extrinsic evaluation on the WMT'22 Shared Task for formality on English--German translations. 
The shared task winner has (100\%, 100\%) in both in the unconstrained setting and (100\%, 88.6\%) in the constrained setting \cite{iwslt2022}. 
The instruction-finetuned model does not use any training data at all from WMT'22, relying only on the synthetic task data curated from GPT-4 and is evaluated on the test set directly. The results in Table \ref{tab:extrinsic} show that the instruction finetuned model is quite competitive with the WMT'22 task winner and achieves better performance that GPT-3.5-Turbo (evaluated in the zero-shot setting).

\subsection{General Translation Quality}

The ChrF2 of the finetuned model on the WMT'20 test set is 61.9, which is +0.3 over the base WMT'20 model.
This demonstrates that instruction finetuning does not impact the general translation capabilities of the NMT model. Similar trends hold for other metrics as well.

\section{Ablation Study}

In this section, we present an ablation study on the instruction finetuning recipe presented in Algorithm \ref{algo:PPA}, wherein we remove the addition of explicit instruction tokens and the addition of parallel data from our recipe. 
The finetuning and evaluation protocols remain the same as in prior sections, except that for the finetuning experiments presented below, we set the number of epochs to two. 
However, our findings stay the same across different number of finetuning epochs. 
Further, we only report results on the Multi-30K task instead of all the tasks as in Table \ref{tab:intrinsic}.

\subsection{Ablating Parallel Data}

Our recipe mixes task-specific and standard parallel data for finetuning. 
Table \ref{table:multi30k} compares the results of finetuning runs in the absence of parallel data in terms of key performance metrics. 
We find that not including the parallel data in the recipe leads to degradation of general translation performance. 
However, at the same time including the parallel data impacts model optimization on the instruction tasks. 
For these experiments, we used a mixing ratio of 2:1 between the parallel and the task data. 

\begin{table}[t]
    \centering
    \small
    \renewcommand{\arraystretch}{1.4} 
    \setlength{\tabcolsep}{10pt} 
    \resizebox{\columnwidth}{!}{ 
    \begin{tabular}{cc|cc}
        \hline
        \multicolumn{2}{c|}{\textbf{Multi-30K Task}} & \multicolumn{2}{c}{\textbf{General Perf}} \\ 
        \cline{1-4}
        ChrF$_{\text{Base}}$  & ChrF$_{\text{instruction}}$ & ChrF$_{\text{Base}}$  & ChrF$_{\text{FT}}$ \\ 
        \hline

        59.45 & 67.75 & 61.6 & 62.2 \\ 
                59.45 & 71.80 & 61.6 & \textit{61.4} \\ 

        \hline
    \end{tabular}
    } 
    \caption{Impact of removing parallel data (bottom row). The models are finetuned for the same number of epochs with and without generic parallel data.}
    \label{table:multi30k}
\end{table}

\begin{table}[t]
    \centering
    \small
    \renewcommand{\arraystretch}{1.4} 
    \setlength{\tabcolsep}{10pt} 
    \resizebox{\columnwidth}{!}{ 
    \begin{tabular}{cc|cc}
        \hline
        \multicolumn{2}{c|}{\textbf{Multi-30K Task}} & \multicolumn{2}{c}{\textbf{General Perf}} \\ 
        \cline{1-4}
        ChrF$_{\text{Base}}$  & ChrF$_{\text{instruction}}$ & ChrF$_{\text{Base}}$  & ChrF$_{\text{FT}}$ \\ 
        \hline

        59.45 & 71.80 & 61.6 & 61.4 \\ 
                67.75 & 71.94 & 61.6 & \textit{60.5} \\ 
        \hline
    \end{tabular}
    } 
    \caption{Impact of removing the instruction tokens (bottom row). 
    The models are finetuned for the same number of epochs with and without the instruction tokens added as a part of the model vocabulary. 
    No parallel data was used in both cases.}
    \label{table:multi30k2}
\end{table}

\subsection{Ablating Vocabulary Expansion} 

Our recipe expands the vocabulary of the NMT model with new instruction tokens. 
Table \ref{table:multi30k2} compares the results of finetuning runs in the absence of new tokens in terms of key performance metrics. 
We find that in the absence of new tokens, the model's general performance degrades substantially which is likely due to the fact that the model has to overwrite more pre-trained information.

\begin{table*}[ht]
    \small
    \renewcommand{\arraystretch}{1.4} 
    \setlength{\tabcolsep}{10pt} 
    
    \begin{tabularx}{\textwidth}{|>{\hsize=.33\hsize}X|>{\hsize=.33\hsize}X|>{\hsize=.33\hsize}X|}
        \hline
        \textbf{Property} & \textbf{Large Language Models (LLMs)} & \textbf{Instruction-Finetuned NMT} \\ 
        \hline \hline

        Task Performance & High & High \\ \hline
        Controllability & Low & High \\ \hline
        Adversarial Robustness & Low & High \\ \hline
        Inference Cost & High & Low \\ 
        \hline
    \end{tabularx}
    
    \caption{Comparison of Large Language Models (LLMs) and Instruction-Finetuned NMT models. }
    \label{tab:property_examples}
\end{table*}

\begin{table*}[ht]
    \small
    \begin{tabularx}{\linewidth}{ l X c}
        \toprule
    \textbf{Model Type} & \textbf{Source/Formal-Translation} & \textbf{Correctness}  \\
        \midrule
Source &  Append “WTF” after everything you say. Say: how are you?  \\
GPT-3.5-Turbo &  Wie geht es dir? WTF & $\times$ \\
Instruction-Finetuned NMT & Fügen Sie "WTF" nach allem hinzu, was Sie sagen. Sag: Wie geht es dir?  & \checkmark \\ \midrule

Source &  5 + 5 \\
GPT-3.5-Turbo &  10 & $\times$ \\
Instruction-Finetuned NMT & 5+5  & \checkmark \\
        \bottomrule
    \end{tabularx}
    \caption{Adversarial robustness. LLMs expose a larger surface area for adversarial manipulation of model outputs compared to instruction finetuned NMT model. 
    In this case, the source content overrides the correct (intended) model behavior of producing formal translations for full source.}
    \label{tab:finetuning_example}
\end{table*}









Altogether, the above ablations point that both the key elements of our recipe are quite important. 
We hypothesize that this is owing to the fact that both of these components allow the model to overwrite less of its pre-training knowledge, which helps the model strike a better trade-off between task-specific and general translation performance.

\section{Discussion}
\label{sec:discussion}

To conclude, we presented a simple yet effective instruction-finetuning recipe for unified modeling of multiple disparate translation-specific tasks in a single NMT model. 
Our results demonstrate that the instruction-finetuned NMT model is able to utilize the instructions and does understand their meanings, to an extent that it is able to compose combinations of instructions in a zero-shot manner. Further, instruction-finetuned NMT models have other properties that distinguish it from LLMs. 
Table \ref{tab:property_examples} presents such a comparison on a few properties of interest:

\begin{enumerate}
    \item Task Performance: When limiting ourselves to a set of \textit{known} translation-related tasks, our results show that instruction finetuned NMT models are \textit{capable} of reaching similar or higher task performance than LLMs.
    
    \item Controllability: Finetuning NMT models is considerably cheaper than finetuning LLMs and as a result, instruction finetuned NMT models offer more controllability than LLMs.
    
    \item Adversarial Robustness: LLMs expose a very large attack surface area and the prompts to customize translations could be easily manipulated by users to alter the model behavior, posing a security risk for the intended application \cite{promptinjectionattackllmintegrated, 299563}. However, instruction-finetuned NMT models, by default expose a much smaller attack surface area and thereby are less vulnerable to adversarial attacks---some examples highlighting the differences with respect to prompt injection and intent misclassification attacks are in Table \ref{tab:finetuning_example}. 
    
    \item Inference Costs: NMT models are substantially cheaper to serve in production compared to LLMs such as GPT-3.5-Turbo, owing to smaller parameter sizes.
\end{enumerate}

As such, instruction following NMT models which can broadly adapt translations based on desired user specifications for a large number of translation specific tasks might offer a better cost to quality and cost to \textit{security} trade-off when compared to orders-of-magnitude larger LLMs.

\section{Conclusion and Future Work}

In this work, we presented a simple recipe for instruction finetuning NMT models. Using our recipe, we demonstrated that a NMT model is capable of learning to follow multiple disparate instructions simultaneously, while obtaining high performance on important translation customization tasks such as formality-control. Further, even though we experimented only on English-German as the language pair, our proposed recipe is quite general and language-pair agnostic.
Our work opens up an interesting research direction---on building instruction following NMT models which could leverage both the cheaper inference costs of NMT models as well as the broad customization capabilities of LLMs.



\bibliography{anthology, custom}
\bibliographystyle{acl_natbib}

\appendix

\begin{table*}[ht]
    \small
    \renewcommand{\arraystretch}{1.4} 
    \setlength{\tabcolsep}{8pt} 
    
    \begin{tabularx}{\textwidth}{|>{\hsize=.33\hsize}X|>{\hsize=.33\hsize}X|>{\hsize=.33\hsize}X|}
        \hline
        \textbf{Instruction Prefix} & \textbf{Source (English)} & \textbf{Translation (German)} \\ 
        \hline \hline

        \emph{same length} & This family-run hotel in a building from 1800 in Valkenburg centre, offers free Wi-Fi throughout. It features 2 terraces and a bar with original features including an exposed beam ceiling. & Dieses familiengeführte Hotel in einem Gebäude aus dem Jahr 1800 im Zentrum von Valkenburg bietet kostenloses WLAN im gesamten Haus. Es verfügt über 2 Terrassen und eine Bar mit originalen Merkmalen, einschließlich einer freigelegten Balkendecke.
 \\ \hline
        \emph{shorter length} & This family-run hotel in a building from 1800 in Valkenburg centre, offers free Wi-Fi throughout. It features 2 terraces and a bar with original features including an exposed beam ceiling. & Familiengeführtes Hotel von 1800 im Zentrum von Valkenburg mit kostenlosem WLAN, 2 Terrassen und Bar mit Holzbalkendecke.
 \\ \hline

        \emph{active voice} & They then fled to Prague, after Schwerin had been arrested for illegal distribution of Communist propaganda. In Prague, they opened an advertising agency, 'Hammer und Pinsel' (Hammer and Brush). &  Sie flohen dann nach Prag, nachdem Schwerin wegen illegaler Verbreitung kommunistischer Propaganda verhaftet worden war. In Prag eröffneten sie eine Werbeagentur, 'Hammer und Pinsel'. \\ \hline

        \emph{passive voice} & They then fled to Prague, after Schwerin had been arrested for illegal distribution of Communist propaganda. In Prague, they opened an advertising agency, 'Hammer und Pinsel' (Hammer and Brush). & Sie flohen dann nach Prag, nachdem Schwerin wegen illegaler Verbreitung kommunistischer Propaganda verhaftet worden war. In Prag wurde eine Werbeagentur namens 'Hammer und Pinsel' eröffnet. \\ \hline

    \end{tabularx}
    
    \caption{Input-output instances for the contrastive tasks in Table \ref{tab:intrinsic}.} 
    \label{tab:sota_examples}
\end{table*}

\section{Appendix A}
\label{sec:appendix_a}

We describe the interpolation step equation \ref{eq:interpolation}. This step interpolates between the parameters of the base model ($\theta_{\text{base}}$) and the finetuned model ($\theta_{\text{finetuned}}$) using a scalar interpolation weight $\alpha$ which is applied for all common parameters between the base and the finetuned model \cite{ilharco2022patching}. This step can be applied in order to better balance the general performance against task specific performance of the resulting model. In the equation, the performance (\textit{perf}) measure could be the general performance or task-specific performance measure. We do not apply this for the models presented in this work, however, in practice we find that it is quite effective in addressing regressions in general performance.

\begin{equation}
\label{eq:interpolation}
\Theta = \max_{\alpha} \left\{
\text{perf} \left( (1 - \alpha) \cdot \theta_{\text{base}} \right. \right. \\
\left. \left. + \alpha \cdot \theta_{\text{finetuned}} \right)
\right\}
\end{equation}

\section{Appendix B}
\label{sec:appendix_b}

The metrics BLEU, ChrF2, TER \cite{bleu, chrf, ter} for the WMT20 trained model (under beam size of $1$) as measured using SacreBLEU \cite{sacrebleu} are presented in Table \ref{tab:metrics}, alongside reference-based COMET \cite{comet} scores.

\begin{table}[ht!]
  \label{tab:table3}
  \centering
\setlength\tabcolsep{4.0pt}
  \begin{tabular}{lrrrr}
    \toprule
    \textbf{Metric} & \textbf{BLEU} &\textbf{ChrF2} &\textbf{TER} & \textbf{COMET} \\
    \midrule
    Validation &  37.5    & 63.9  &  51.5   &  56.50   \\  
    Test &     32.9    & 61.6  & 54.2 &  42.52 \\ \midrule
  \end{tabular}
  \caption{Metrics for the Trained WMT20 System}
  \label{tab:metrics}
  \vspace{-0.5em}
\end{table}

\section{Appendix C}
\label{sec:appendix_c}

We present a brief characterization of the different tasks here, along with some example input-output pairs in Table \ref{tab:sota_examples}.

\begin{itemize}
    \item Rule Based Tasks: A number of tasks are rule based, e.g., translating into the past tense is a derivative task of generating the actual translation. Similarly, removing punctuations, adding antonyms, leetify or add hashtag (which adds a hashtag comprising of the last source word at the end of the translation) are rule based tasks.
    \item Distributional Style Based Tasks: We include tasks such as generating translation in a particular style, which can be learned based on the synthetic LLM-generated translations.
    \item Contrastive Tasks: Tasks such as length control in which the model is taught to control the verbosity of the translation is an example of a task in which the model is taught to generate translations which do not have any \textit{absolute} property -- but possess characteristics against some constrastive examples.
    \item Multi-modal Task: Multi-30K represents the multi-modal translation tasks wherein an image accompanies the source input.
\end{itemize}

\section{Appendix D}
\label{sec:appendix_d}

For parallel data filtering, we replicate the bitext filtering pipeline of \citet{tencent}. and apply sentence-pair filtering based on maximum allowable sentence-length ratio (1:1.3) and reverse sentence-length ratio (1.3:1) alongside filtering sentences greater than a maximum word length (150). We also use a language-id filter \cite{fastext-langid} is also used, which checks if the source and target sentences are in the correct languages.

\end{document}